\documentclass[conference]{IEEEtran}
\IEEEoverridecommandlockouts
\usepackage{cite}
\usepackage{amsmath,amssymb,amsfonts}
\usepackage{algorithmic}
\usepackage{graphicx}
\usepackage{textcomp}
\usepackage{xcolor}
\usepackage[ruled,linesnumbered]{algorithm2e}
\usepackage{booktabs}
\usepackage{multirow}
\def\BibTeX{{\rm B\kern-.05em{\sc i\kern-.025em b}\kern-.08em
    T\kern-.1667em\lower.7ex\hbox{E}\kern-.125emX}}
\begin{document}

\title{GapDNER: A Gap-Aware Grid Tagging Model for Discontinuous Named Entity Recognition\\
}

\author{\IEEEauthorblockN{1\textsuperscript{st} Yawen Yang}
\IEEEauthorblockA{\textit{School of Software} \\
\textit{Tsinghua University}\\
Beijing, China \\
yyw19@mails.tsinghua.edu.cn}
\and
\IEEEauthorblockN{2\textsuperscript{nd} Fukun Ma}
\IEEEauthorblockA{\textit{School of Software} \\
\textit{Tsinghua University}\\
Beijing, China \\
mfk22@mails.tsinghua.edu.cn}
\and
\IEEEauthorblockN{3\textsuperscript{rd} Shiao Meng}
\IEEEauthorblockA{\textit{School of Software} \\
\textit{Tsinghua University}\\
Beijing, China \\
\qquad msa21@mails.tsinghua.edu.cn \qquad}
\and
\IEEEauthorblockN{4\textsuperscript{th} Aiwei Liu}
\IEEEauthorblockA{\textit{School of Software} \\
\textit{Tsinghua University}\\
Beijing, China \\
liuaw20@mails.tsinghua.edu.cn}
\and
\IEEEauthorblockN{5\textsuperscript{th} Lijie Wen$^{*}$\thanks{$^{*}$Corresponding author.}}
\IEEEauthorblockA{\textit{School of Software} \\
\textit{Tsinghua University}\\
Beijing, China \\
wenlj@tsinghua.edu.cn}
}
\maketitle

\begin{abstract}
In biomedical fields, one named entity may consist of a series of non-adjacent tokens and overlap with other entities. Previous methods recognize discontinuous entities by connecting entity fragments or internal tokens, which face challenges of error propagation and decoding ambiguity due to the wide variety of span or word combinations. To address these issues, we deeply explore discontinuous entity structures and propose an effective Gap-aware grid tagging model for Discontinuous Named Entity Recognition, named GapDNER. Our GapDNER innovatively applies representation learning on the context gaps between entity fragments to resolve decoding ambiguity and enhance discontinuous NER performance. Specifically, we treat the context gap as an additional type of span and convert span classification into a token-pair grid tagging task. Subsequently, we design two interactive components to comprehensively model token-pair grid features from both intra- and inter-span perspectives. The intra-span regularity extraction module employs the biaffine mechanism along with linear attention to capture the internal regularity of each span, while the inter-span relation enhancement module utilizes criss-cross attention to obtain semantic relations among different spans. At the inference stage of entity decoding, we assign a directed edge to each entity fragment and context gap, then use the BFS algorithm to search for all valid paths from the head to tail of grids with entity tags. Experimental results on three datasets demonstrate that our GapDNER achieves new state-of-the-art performance on discontinuous NER and exhibits remarkable advantages in recognizing complex entity structures.

\end{abstract}


\section{Introduction}
Named entity recognition (NER) aims to identify mentions with specific meanings in text and classify them into pre-defined entity categories, including person, location, gene, disease, etc. NER has long been a fundamental task in natural language processing because of its wide applications in downstream tasks like entity linking \cite{b1}, relation extraction \cite{b2} and knowledge graph \cite{b3}. 

\begin{figure}[htbp]
\centerline{\includegraphics[width=0.95\linewidth]{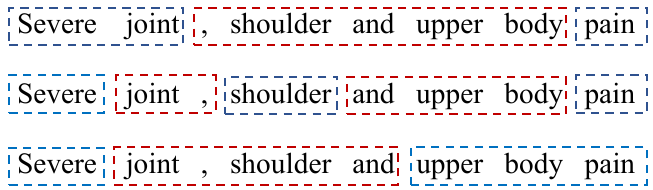}}
\caption{Example of three discontinuous entities (``\emph{severe joint pain}'', ``\emph{severe shoulder pain}'' and ``\emph{severe upper body pain}'') within the same sentence. The blue and red dashed boxes represent the fragments of discontinuous entities and context gaps respectively.}
\label{fig1}
\end{figure}

Traditional entity definition contains two underlying assumptions that an entity should be composed of consecutive adjacent tokens and should not overlap with other ones \cite{b4}. While in practical applications, there are often complex entity structures that do not conform to the above assumptions, including nested, overlapped and discontinuous entities. Among them, the discontinuous entity denotes an entity consisting of a sequence of non-adjacent tokens. As illustrated in Fig.~\ref{fig1}, ``\emph{severe shoulder pain}'' represents a discontinuous entity separated by two context gaps. Since typical BIO tagging methods fail to handle discontinuous entities with diverse overlapped structures, discontinuous NER still remains a challenge for most NER systems.

Some meaningful efforts have been devoted to explorations of discontinuous NER. Existing methods could be primarily divided into sequence-labeling, span-based, seq2seq-based and grid-tagging solutions. \textbf{1) Sequence labeling methods} extend the common BIO tagging mode to more complex tagging schemes such as BIOHD \cite{b5} and the 10-tag scheme \cite{b6}. Such task-specific extensions restrict the adaptability of NER models to new entity types and unseen data sets. \textbf{2) Span-based models} \cite{b7} recognize discontinuous entities by identifying entity fragments and combining them together, which are limited by maximum span length and struggle in decoding ambiguity. \textbf{3) Seq2seq-based methods} \cite{b8},\cite{b9},\cite{b10} transform discontinuous entity recognition into entity span sequence generation task and adopt the pre-trained seq2seq model to solve it. These methods have no need of span connections, thus avoiding the problem of decoding ambiguity. However, they suffer from error accumulation and low inference efficiency caused by the autoregressive mechanism. \textbf{4) Grid tagging models} \cite{b11},\cite{b12},\cite{b13}, which formulate entity extraction as token-pair grid classification, have achieved excellent performance in discontinuous NER. But they overlook the semantic relations between different spans and still suffer from decoding ambiguity problem during token connection. Most recently, \textbf{large language models (LLMs)} have been explored to offer general solutions to entity recognition through prompt engineering and instruction tuning. But their performances \cite{b14,b15} on discontinuous NER are still far from current state-of-the-art models. LLMs, which are optimized for next token prediction, are not satisfactory entity extractors currently.

To reduce the above drawbacks, we explore deep into boundaries and structures of discontinuous entities. From Fig.~\ref{fig1}, we discover that each range of discontinuous entities follows the rule that entity fragments and context gaps are arranged alternatively. Consequently, identifying context gaps correctly will certainly help locate entity fragments and alleviate decoding ambiguity. Additionally, entity fragments provide important clues for entity type recognition, and context gaps always contain non-entity tokens. It is beneficial to capture the internal regularity of each span before performing span classification. Furthermore, discontinuous entities are often accompanied by overlapped situations. Learning the semantic relations between different spans may further improve the classification performance.

Armed with these observations, we propose an effective \textbf{Gap}-aware grid tagging model for \textbf{D}iscontinuous \textbf{NER}, called \textbf{GapDNER}, to fully utilize gap information. We regard context gaps as special spans and convert span classification into the token-pair grid tagging problem. Next, we design two sequential modules to model semantic features from both intra- and inter-span aspects. The intra-span regularity extraction module empirically employs the Biaffine mechanism on token pairs and skillfully applies linear attention to all tokens within each token pair, aiming to capture the internal regularity of each span (namely entity fragment or context gap). After that, we concatenate the two representations to construct the token-pair feature matrix. We notice that in the overlapped structures of discontinuous entities, several spans share the same head or tail token and thus appear in the same row or column of the token-pair grid. Therefore, the inter-span attention enhancement module calculates the criss-cross attention over the whole feature matrix to measure inter-span relations and facilitate span classification. At the inference stage, we allocate a directed edge to each span, then employ the BFS (Breadth-First Search) algorithm to search for all valid paths from the head to the tail of token pairs with entity type labels. Each path corresponds to an actual entity. The main contributions of this work can be listed as follows:
\begin{enumerate}
\item[$\bullet$] We propose an effective gap-aware grid tagging model for discontinuous NER, which innovatively leverages context gap information to enhance entity recognition. To the best of our knowledge, we are the first to apply the representation learning on context gaps in explorations of discontinuous NER.
\item[$\bullet$] We adopt Biaffine mechanism along with linear attention to obtain the intra-span regularity and employ the criss-cross attention to capture the inter-span relations. The integration of intra- and inter-span features significantly promotes span classification. 
\item[$\bullet$] We conduct comprehensive experiments on three standard benchmarks for discontinuous NER. Experimental results show that our GapDNER outperforms strong baselines consistently on each dataset and achieves new state-of-the-art results in diverse settings.
\end{enumerate}

\section{Related Work}
\subsection{Discontinuous NER}
Discontinuous NER (DNER) involves recognizing entities composed of a sequence of non-adjacent tokens, which is of great significance for medical record recognition and drug safety monitoring. Existing methods for DNER can be mainly divided into the following categories.

\textbf{Sequence labeling} methods assign a label to each token that indicates the location and type of entities. Tang et al. \cite{b5} extended the BIO scheme to BIOHD by introducing two additional tags to distinguish overlapping entity fragments. Corro \cite{b6} proposed a new tagging scheme consisting of 10 labels to mark tokens of diverse complex structures. Although effective to some extent, these methods have limitations of tagging inflexibility and decoding ambiguity.

\textbf{Span-based} methods always transform discontinuous NER into a two-stage process that includes span detection and span connection. Li et al. \cite{b7} developed a span-based model to jointly recognize overlapped and discontinuous entities by enumerating all possible spans and identifying the neighboring relation between them. Due to the enumeration nature, these methods are limited by the maximum span length and have higher complexity when recognizing long entities.

\textbf{Seq2seq-based} methods uniformly formulate entity recognition as entity span sequence generation. BartNER \cite{b8}, which incorporated the pre-trained seq2seq model BART and a pointer network, was the first to employ the generative framework to address various NER subtasks. After that, Debias-DDA \cite{b9} and Debias-CSL \cite{b10} reduced optimization objective bias and fixed-order bias of seq2seq models, respectively. Unfortunately, they face challenges of error propagation and low inference efficiency caused by the autoregressive mechanism.

\textbf{Grid tagging} methods have recently aroused increasing attention by converting entity identification into token-pair grid classification. Wang et al. \cite{b11} reformulated DNER as maximal clique discovery and first proposed grid tagging solutions. Another typical approach is W2NER \cite{b12}, which proposed to model multi-type NER as the word-word relation classification. Inspired by that, Liu et al. \cite{b13} put forward another grid tagging model for DNER by introducing two additional relation tags. Although these models have achieved excellent performance, they still suffer from decoding ambiguity when handling complex overlapped structures.

Other approaches for DNER include the LLM-related \cite{b14, b15}, hypergraph-based \cite{b16} and transition-based \cite{b17} models.

\subsection{Optimizations of Grid Attention Computation}
In computer vision, it is essential to perform attention calculations on the 2D grid of feature maps. However, capturing global attention offen results in excessive computational complexity. To address this difficulty, several optimization methods have been proposed for approximate computation.

\textbf{Axial attention} \cite{b18} calculates attention scores along a specific axis (horizontal or vertical) of the feature map, mixing feature information from that axis while keeping information along other axes independent. Most related works combine two axial attention layers of different axes to make the neural model acquire global receptive fields. \textbf{Criss-cross attention} \cite{b19} fuses feature information in a criss-cross manner, which is similar to the combination of row attention and column attention. It enables each pixel not only to pay attention to the local area but also to transmit information across regions, breaking the limitations of traditional local receptive fields. \textbf{Dilated attention} \cite{b20}, soured from the dilated convolution, introduces a dilation factor to expand the receptive field when calculating attentions on a feature map. It makes grid elements attend to information at a greater distance without increasing the computational complexity.

In this work, we employ the criss-cross attention to model inter-span relations due to the complex overlapped structures among discontinuous entities.

\section{Gap-Aware Grid Tagging Scheme}
We regard context gaps as special mentions and define four types of span used for entity recognition. \textbf{1) ConE} denotes a continuous entity. \textbf{2) DiscE} denotes the range of a discontinuous entity (e.g., from ``\emph{severe}'' to ``\emph{pain}'' in Fig.~\ref{fig1}). \textbf{3) Frag} represents a continuous fragment of one discontinuous entity (e.g., ``\emph{severe joint}'' in Fig.~\ref{fig1}). \textbf{4) Gap} represents a context interval between two neighboring fragments of the same discontinuous entity (e.g., ``\emph{and upper body}'' in Fig.~\ref{fig1}).

Equipped with these definitions, we predefine the token pair grid label set $C$ for both continuous and discontinuous entities. Considering one span can be both a continuous entity and a fragment of another discontinuous entity simultaneously, we annotate the (head, tail) grid with the same fragment label (``$\langle \text{Frag} \rangle$'') to uniformly represent them. Meanwhile, we distribute entity type labels (e.g., ``$\langle \text{ADE} \rangle$'', ``$\langle \text{Disorder} \rangle$'') to the (tail, head) grid of each ConE or DisE span, which serve as the begin flag for entity decoding. Finally, we annotate the (head, tail) grid of a context gap with the ``$\langle \text{Gap} \rangle$'' label. Such operations face a conflict in situations where one token is both an entity and a fragment. To solve this, we choose to annotate the one-word entity with the corresponding entity type regardless of whether it is a fragment, because each one-word entity will be definitely considered as a potential fragment in the inference stage.

\begin{figure}[htbp]
\centerline{\includegraphics[width=0.95\linewidth]{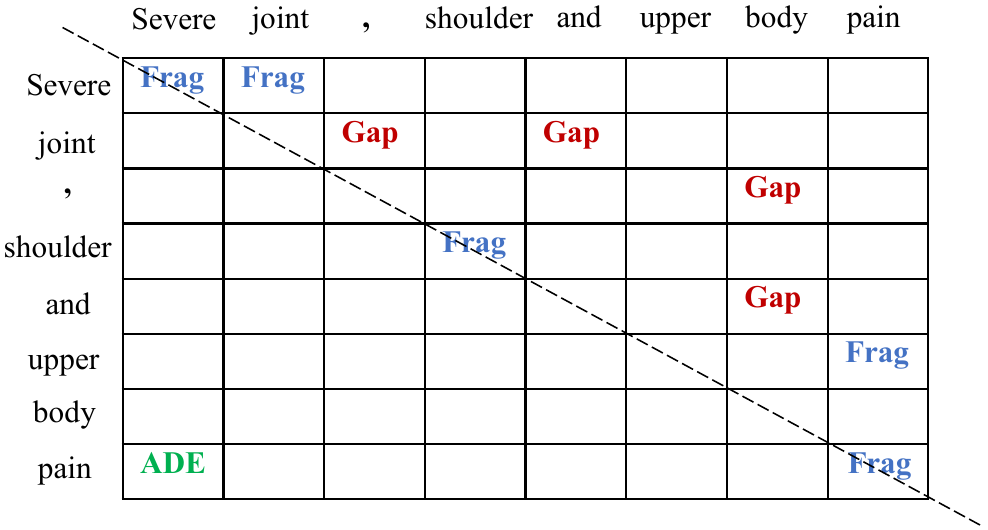}}
\caption{An example of our gap-aware grid tagging scheme. Blue and red fonts represent fragment and gap labels respectively, while green fonts denote entity type labels. The black dotted line covers the diagonal area of the grid.}
\label{fig2}
\vspace{-2.7mm}
\end{figure}

Under this tagging scheme, the upper triangular region of the grid can only include ``$\langle \text{Frag} \rangle$'' and ``$\langle \text{Gap} \rangle$'' labels. The lower triangle region can only contain entity type labels like ``$\langle \text{ADE} \rangle$''. Specially, the diagonal positions can contain all the above types since they stand for spans with only one token. For better understanding, we provide an example of grid tagging scheme in Fig.~\ref{fig2}, corresponding to the input sentence.

\section{Proposed Model}

\begin{figure*}[htbp]
\centerline{\includegraphics[width=\linewidth]{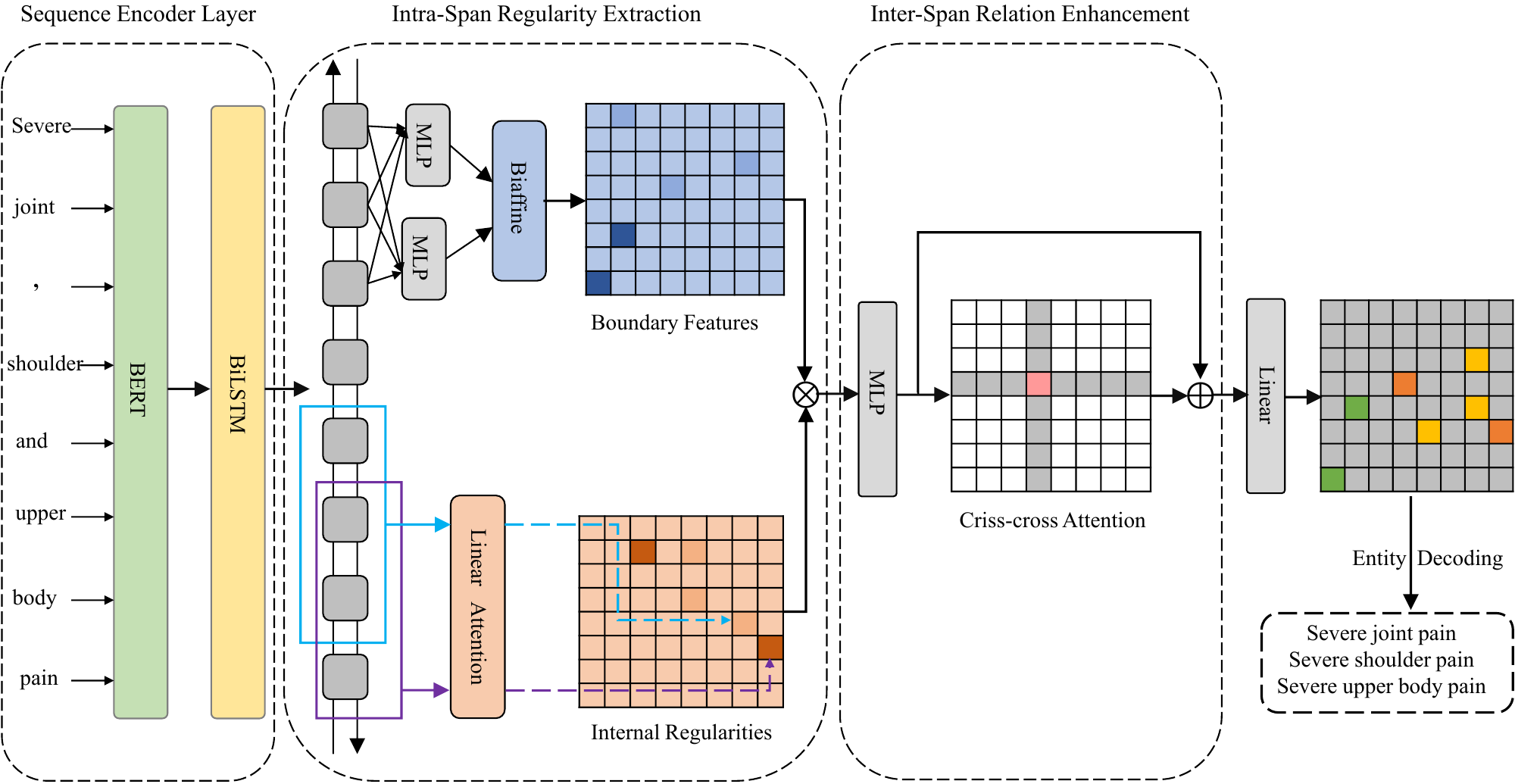}}
\caption{The overall architecture of proposed GapDNER model. $\otimes$ denotes vector concatenation of the last dimension. $\oplus$ represents the element-wise addition.}
\label{fig3}
\vspace{-3mm}
\end{figure*}

The overall model architecture is illustrated in Fig.~\ref{fig3}, which mainly includes three components: Sequence Encoder Layer, Intra-Span Regularity Extraction and Inter-Span Relation Enhancement. During model inference, we decode entities based on classification results of the token-pair feature grid.

\subsection{Sequence Encoder Layer}\label{AA}
We adopt BERT \cite{b21} and BiLSTM \cite{b22} to construct sequence encoder layer. The pre-trained model BERT has been demonstrated to be one of the outstanding models for representation learning in information extraction. BiLSTM excels at modeling sequence order and local dependencies. Given an input sentence of $n$ tokens $s=[t_1, t_2, ..., t_n]$, we feed it into the BERT network to obtain context-dependent embeddings. After the BERT calculation, one target token may be divided into several word pieces according to BERT dictionary. Following pervious works \cite{b12}, we employ the max pooling operation to generate token representation from word piece embeddings. Then the sequence of token representations is put into BiLSTM module to produce the final token representation $\mathbf{H}=[h_1, h_2, ..., h_n] \in \mathbb R^{n \times d}$, which can be formulated as:
\begin{align}
x_i &= \text{BERT}(t_i), \\
\overrightarrow{h}_i &= \text{LSTM}_f(\overrightarrow{h}_{i-1}, x_i;\theta_f), \\
\overleftarrow{h}_i &= \text{LSTM}_b(\overleftarrow{h}_{i-1}, x_i;\theta_b), \\
h_i &=  [\overrightarrow{h}_i; \overleftarrow{h}_i],
\end{align}
where $n$ is the sequence length, $d$ is the hidden dimension of token representations after BiLSTM, $[\cdot ; \cdot]$ concatenates two vectors in the last dimension, $\theta_f$ and $\theta_b$ denote the parameters of the forward and backward LSTM, respectively.

\subsection{Intra-Span Regularity Extraction}
In this module, we aim to learn span representations that incorporate internal regularities for better classification. Typical span-based methods for NER represent span features via concatenating head and tail embeddings directly. Subsequent revisions include span length feature integration \cite{b7}, and Biaffine decoder \cite{b23} which strengths the semantic interaction between head and tail tokens. Nevertheless, these methods only capture insufficient and coarse-grained span features since they concentrate mainly on span boundaries but ignore the internal composition regularity.

To effectively learn the semantic regularity of entity fragments and context gaps within discontinuous entities, we employ Biaffine mechanism and linear attention to explore boundary features and internal regularity, respectively. For the former, we first exploit two separate multilayer perceptrons (MLPs) to map the head and tail tokens into different semantic spaces. Then we apply a biaffine decoder over the two representation sequences to generate span boundary feature matrix $\textbf{H}^{(\text{span})} \in \mathbb R^{n \times n \times d}$. Each element is computed as follows:
\begin{align}
\overline{h}_i &= \text{MLP}_\text{head}(h_i), \\
\overline{h}_j &= \text{MLP}_\text{tail}(h_j), \\
h_{ij}^{(\text{span})} &= \overline{h}_i^{\top}U^{(1)}\overline{h}_j + [\overline{h}_i; \overline{h}_j]U^{(2)} + b_1, 
\end{align}
where $\overline{h}_i,\overline{h}_j \in \mathbb R^{d \times 1}$ are the head and tail representations of token pair $(i,j)$ after semantic mapping. $U^{(1)}$ denotes a $d \times d \times d$ tensor, $U^{(2)}$ represents a $2d \times d$ matrix, and $b_1 \in \mathbb R^{1 \times d}$ is the bias. $U^{(1)},U^{(2)},b_1$ are all learnable parameters.

It is worth noting that regularity information stems from each token in the span. The RICON method \cite{b24} has compared the performance of several regularity extracting ways (e.g., Linear attention, Max-pooling, Multi-Head attention) on Chinese NER and discovered linear attention outperformed others. Inspired by that, we select the simple but effective linear attention to extract the intra-span regularity $\textbf{H}^{(\text{reg})} \in \mathbb R^{n \times n \times d}$. For each token pair grid $(i,j)$, assuming $i < j$, the attention score and regularity representation are calculated as:
\begin{align}
a_t &= W_{\text{reg}}^{\top} h_t + b_{\text{reg}}, \\
\alpha_t &= \frac{exp(a_t)}{\sum_{k=i}^j exp(a_k)}, \\
h_{ij}^{(\text{reg})} &= \sum_{t=i}^j \alpha_t \cdot h_t,
\end{align}
where $t \in \{i,i+1,...,j\}$ stands for the token index of the span, $W_{\text{reg}} \in \mathbb R^{d \times 1}$ and $b_{\text{reg}} \in \mathbb R^1$ are learnable weights and bias. Especially, for the span which has only one token (namely $i=j$), we naturally adopt the hidden representation $h_t$ from BiLSTM to indicate its regularity. Since entity type labels are located in the lower triangular region of the grid where $i>j$, we introduce another weights vector and bias to compute the linear attention and corresponding regularity representation. Then we concatenate span boundary and internal regularity representations as the token-pair grid feature $\textbf{G} \in \mathbb R^{n \times n \times 2d}$.
\begin{equation}
\textbf{G} = [\textbf{H}^{(\text{span})};\textbf{H}^{(\text{reg})}]
\end{equation}

\subsection{Inter-Span Relation Enhancement}
As mentioned above, different spans have rich spatial and semantic relations such as neighboring, nested, overlapped and head-tail. It should be beneficial to span identification if we are capable of leveraging these special correlations. To achieve this goal, we calculate the inter-span attention over the token-pair grid and obtain attention-enhanced grid representations for subsequent predictions.

We observe that discontinuous entities often appear together with overlapped structures in biomedical documents. Such co-occurrence situations possibly make entity fragments or context gaps of different entities share the same head or tail token. 
Take Fig.~\ref{fig2} for example, ``\emph{severe}'' and ``\emph{severe joint}'' are both entity fragments sharing the same head boundary, thus they are located in the same row of the grid. Similarly, ``\emph{, shoulder and upper body}'' and ``\emph{and upper body}'' denote two context gaps containing the same tail token, and are distributed in the same column of the token-pair grid.

Considering the obvious axial distribution characteristics of different spans, we introduce criss-cross attention \cite{b19} to measure the inter-span relations and reduce computational complexity. As depicted in Fig.~\ref{fig3}, the criss-cross attention module collects interactive information in horizontal and vertical directions to enhance span-level representative capability. To be specific, we first utilize one MLP to map the concatenated span regularity features into the same semantic space, and reduce vector dimension simultaneously.
\begin{equation}
\textbf{M}=\text{MLP}(\textbf{G})
\end{equation}
where $\textbf{M} \in R^{n \times n \times d}$. Next, the attention module adopts two convolutional layers with $1\times1$ filters on \textbf{M} to generate the query and key matrix \textbf{Q} and \textbf{K} respectively, where $\textbf{Q}, \textbf{K} \in \mathbb R^{n \times n \times d^\prime}$. Another convolutional layer with $1\times1$ filters is applied on \textbf{M} to produce the value matrix \textbf{V}$\in \mathbb R^{n \times n \times d}$. For each token pair grid $q_{ij}$ of \textbf{Q}, we obtain the corresponding criss-cross set $\textbf{K}_{\textbf{cc}} \in \mathbb R^{(n+n-1) \times d^\prime}$, $\textbf{V}_{\textbf{cc}} \in \mathbb R^{(n+n-1) \times d}$ by extracting feature vectors which are in the same row or column with grid $q_{ij}$ from \textbf{K} and \textbf{V}. Particularly, for the grid position $(i,j)$, the corresponding component $m_{ij}^\prime$ of the criss-cross attention output $\textbf{M}^\prime \in \mathbb R^{n \times n \times d}$ is defined as follows:
\begin{align}
m_{ij}^\prime &= \text{CC\_Attention}(q_{ij},\textbf{K}_{cc},\textbf{V}_{cc}) \nonumber \\
       &= \text{Softmax}(\frac{q_{ij}\textbf{K}_{cc}^{\top}}{\sqrt{d^\prime}})\textbf{V}_{cc}, \ 1 \leq i,j \leq n,
\end{align}
where $n$ is the length of input sentence and also denotes the height and width of the token pair grid. $\textbf{K}_{\textbf{cc}}$ and $\textbf{V}_{\textbf{cc}}$ represent keys and values selected from the feature vectors \textbf{K} and \textbf{V}. Given the query $q_{ij}$ located in $(i,j)$, keys and values positioned at the union set of same row and column elements (formally $\{(i^\prime,j^\prime)|i^\prime=i\ \text{or}\ j^\prime=j\}$) will be selected to perform attention computation. In this way, the proposed inter-span relation enhancement module can model the overlapped and neighboring relations between multiple spans, and can capture long-range dependency effectively.

After that, we conduct the element-wise addition to integrate the attention output with initial feature matrix.
\begin{equation}
\textbf{M}^{\prime\prime} = \textbf{M}^\prime + \textbf{M},
\end{equation}
where $\textbf{M}^{\prime\prime} \in \mathbb R^{n \times n \times d}$ denotes the relation enhanced output.

\subsection{Model Training}
Based on the grid representation enhanced by inter-span relations, we exploit a linear layer to obtain the classification logits and adopt the softmax function to calculate the label probability $\hat{y}_{ij}$ for the token pair $(t_i, t_j)$ as:
\begin{equation}
\hat{y}_{ij} = \text{Softmax}(\text{Linear}(\textbf{M}_{ij}^{\prime\prime})),
\end{equation}
where $y_{ij} \in \mathbb R^{|C|}$ denotes the probability vector of the grid tags predefined in $C$. For each sentence $s=[t_1,t_2,...,t_n]$, our training target is to minimize the negative log-likelihood (NLL) loss between the predicted probability and the corresponding gold labels, formulated as:
\begin{equation}
\mathcal{L} = -\frac{1}{n^2}\sum_{i=1}^{n}\sum_{j=1}^{n}\sum_{c=1}^{|C|}y_{ij}^clog\hat{y}_{ij}^c,
\end{equation}
where $n$ is the number of sentence tokens, $y_{ij}$ is the one-hot binary vector with regards to gold class labels, $\hat{y}_{ij}$ represent the predicted probability vector. The superscript $c$ indicates the $c$-th class of the predefined grid tag set $C$.

\subsection{Entity Decoding}\label{SCM}
In the inference stage, the predictions of our model are token pairs and their corresponding grid labels. Following the grid tagging scheme, the entity type label indicates the tail-head token pair of one or more entities belonging to the corresponding category. The ``$\langle \text{Frag} \rangle$'' label represents the head-tail token pair of one continuous entity or fragment of a discontinuous entity. Moreover, the ``$\langle \text{Gap} \rangle$'' tag stands for the head-tail token pair of a context gap between two neighboring fragments of a discontinuous entity.

During entity decoding, we treat the head and tail token of continuous or discontinuous entities as the start and end nodes, assign a directed edge from the head to tail of each fragment and gap span. Then the entity recognition process is converted into the directed path-searching problem. In order to make the end of the previous edge correspond to the beginning of the next edge, we uniformly move the end of each directed edge to the next token. Considering that several discontinuous entities can possibly share the same tail-head token pair, we employ the BFS (Breadth-First Search) algorithm to search for all valid paths, each of which stands for an entity mention. 

\begin{figure}[htbp]
\centerline{\includegraphics[width=\linewidth]{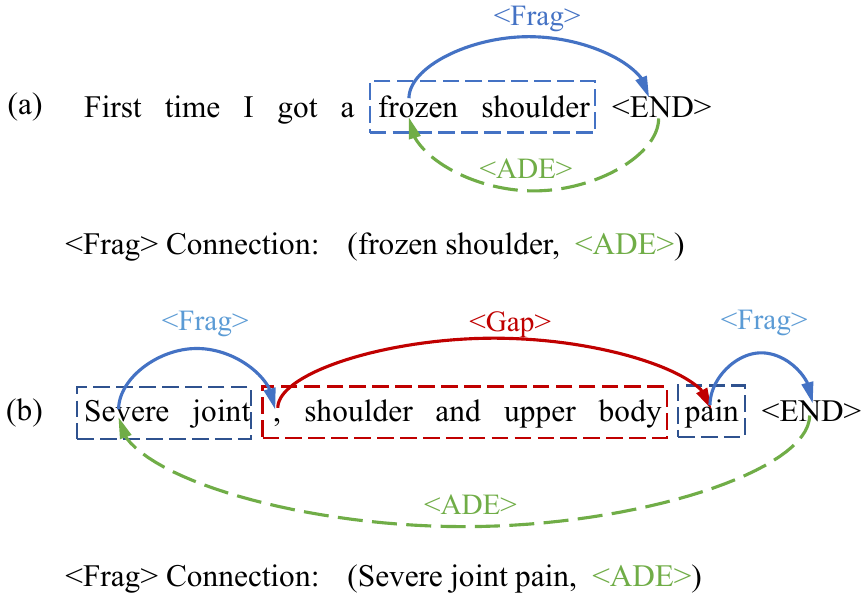}}
\caption{Unified decoding process of continuous and discontinuous entities, including path searching and fragment connection. The end nodes are uniformly moved to the next token to match the start nodes of the next edge.}
\label{fig4}
\vspace{-1.5mm}
\end{figure}

One valid path should meet two requirements: 1) the path contains an odd number of spans and the first edge cannot be a gap; 2) fragments and gaps should be arranged alternately if the path has more than one edge. After path searching operation, we obtain the final entity by connecting entity fragments within each valid path in turn. Fig.~\ref{fig4} shows the decoding process of continuous and discontinuous entities.

\section{Experiments}
\subsection{Datasets}
To evaluate the performance of our proposed model on discontinuous NER, we choose three public NER datasets in the biomedical domain, namely CADEC \cite{b25}, ShARe13 \cite{b26} and ShARe14 \cite{b27}. The CADEC dataset, short for CSIRO Adverse Drug Event Corpus, is a rich annotated corpus of patient-reported adverse drug events (ADEs). Its data are sourced from medical forum posts and contain annotated mentions of symptoms, diseases and ADEs. Following previous researches \cite{b12},\cite{b17}, we only use the adverse drug events (ADE) entities for assessments. ShARe13 and ShARe14 datasets, derived from the Shared Annotated Resources (ShARe) project and the CLEF Initiative, are used for the ShARe/CLEF eHealth Challenge in 2013 and 2014, respectively. They consist of many narrative clinical reports extracted from the MIMIC database, which are annotated for the entity recognition and normalization of disorders.

We utilize the preprocessing scripts provided by Dai et al. \cite{b17} for data cleaning and splitting. Table 1 presents the statistics of each discontinuous NER dataset. It is worth noting that around 10\% entities are discontinuous in each corpus.

\begin{table}
\caption{\label{data statistics}
Data Statistics of three NER datasets.
}
\centering
\resizebox{\linewidth}{!}{ 
\begingroup
\renewcommand{\arraystretch}{0.8} 
\begin{tabular}{l|ccc}
\toprule
& \textbf{CADEC} & \textbf{ShARe13} & \textbf{ShARe14} \\
\midrule
Total documents & 1250 & 299 & 431 \\
Total Sentences & 7597 & 18767 & 34618 \\
Total Entities & 6318 & 11148 & 19073 \\
Entity Type & ADE & Disorder & Disorder \\
\midrule
Disc. Entities & 679 & 1088 & 1658 \\
- Percentage & 10.75\% & 9.76\% & 8.69\% \\
- Number of tokens & 1-13 & 1-7 & 1-7 \\
\bottomrule
\end{tabular}
\endgroup
}
\vspace{-1.5mm}
\end{table}

\subsection{Baselines and Metrics}
We compare the proposed model with baseline methods designed for discontinuous or multi-type NER tasks.\\
$\bullet$ \textbf{Sequence-labeling methods: }\textbf{BIOHD} \cite{b5} extended the BIO tagging scheme to BIOHD mode and combined LSTM and CRF network to identify discontinuous entities by sequence labeling. \textbf{Corro} \cite{b6} was recently proposed to improve the sequence tagging scheme through a two-layer representation of discontinuous entities.\\
$\bullet$ \textbf{Span-based methods: }\textbf{JointNER} \cite{b7} proposed a span-based model which could jointly tackle overlapped and discontinuous entities by learning the neighoring relation between spans. \\
$\bullet$ \textbf{Seq2seq-based methods: }\textbf{BartNER} \cite{b8} developed an effective seq2seq model with the pointer network to solve three types of NER subtasks. \textbf{MAPtr} \cite{b28} proposed a generative model for discontinuous NER based on pointer networks, where the pointer indicated the role of current token and the location of next constituent token. \textbf{Debias-DDA} \cite{b9} and \textbf{Debias-CSL} \cite{b10} explored the incorrect biases in the generation process and further improved the performance through deconfounding data augmentation and calibrating sequence likelihood, respectively.\\
$\bullet$ \textbf{Grid-tagging methods: }\textbf{MAC} \cite{b11} first introduced the grid tagging scheme to predict entity fragments and their connecting relations within a discontinuous entity. \textbf{W2NER} \cite{b12} formulated different NER subtasks into the classification problem of word pair relations and decoded discontinuous entities with grid tags defined as Next-Neighboring-Word (NNW) and Tail-Head-Word (THW). \textbf{TOE} \cite{b13} improved the tagging scheme of W2NER by adding two grid labels, namely Previous-Neighboring-Word (PNW) and Head-Tail-Word (HTW), to obtain more fine-grained relations and alleviate error propagation. \\ 
$\bullet$ \textbf{LLM-related methods: }\textbf{LLaMA2-13B} \cite{b14} developed a cascade instruction tuning method to transfer domain-specific LLMs to biomedical NER tasks. \textbf{LLaMA-7B} \cite{b15} proposed an effective prompt template for LLMs, which incorporates label-injected instructions for few-shot NER applications. \\
$\bullet$ \textbf{Others: }\textbf{SegHyper} \cite{b16} employed segmental hypergraphs to extract all candidate spans, then merged them into discontinuous entities. \textbf{TBM} \cite{b17} was a transition-based end-to-end model with generic neural encoding for discontinuous NER.

For the evaluation metrics, we employ the span-level precision, recall and F1 score, based on the exact matching of entity token indexes and categories.
\subsection{Implementation details}
For each dataset, we predefine the token pair grid label set $C$ with ``None'', ``Frag'', ``Gap'' and each entity type (e.g., ``ADE''). Thus the length of grid label set is $|T|+3$, where $T$ denotes the initial label set of entity categories.

With regard to model components, we employ the pretrained BioBERT \cite{b29} for CADEC, Clinical BERT \cite{b30} for ShARe13 and ShARe14, which is the same as other baselines \cite{b12,b13} for fair comparisons. We use the AdamW optimizer with the learning rate set to 5e-6 for BERT and 1e-3 for other modules in model training. The total epoch number is 15 and the batch size is set to 12 for CADEC, and 8 for ShARe13 and ShARe14. To avoid overfitting, we apply the dropout rate of 0.5 to each dataset. The best-performing model is taken according to the micro-F1 score on the validation data set. For each experiment, we report the average F1 results with 5 runs of training and testing. All experiments are conducted on a single GeForce RTX 3090 GPU.

\begin{table*}
\caption{\label{main-result-disc}
Overall performance on discontinuous NER datasets. $\dagger$ represents our reproduction results via the source codes. \textbf{Bold} denotes the best score while \underline{underline} indicates the second best.
}
\centering
\resizebox{0.90\linewidth}{!}{ 
\begin{tabular}{llccccccccc}
\toprule
\multicolumn{2}{c}{\multirow{2}*{\textbf{Model}}} & \multicolumn{3}{c}{\textbf{CADEC}} & \multicolumn{3}{c}{\textbf{ShARe13}} & \multicolumn{3}{c}{\textbf{ShARe14}} \\
\cmidrule(r){3-5}
\cmidrule(r){6-8}
\cmidrule(r){9-11}
~ & ~ & \textbf{P} & \textbf{R} & \textbf{F1} & \textbf{P} & \textbf{R} & \textbf{F1} & \textbf{P} & \textbf{R} & \textbf{F1} \\
\midrule
\multirow{2}*{\textbf{Sequence-labeling}} & BIOHD \cite{b5}  & 67.80 & 64.99 & 66.36 & - & - & - & - & - & -\\
~ & Corro \cite{b6} & - & - & \underline{72.90} & - & - & 82.10 & - & - & 81.80 \\
\midrule
\textbf{Span-based} & JointNER \cite{b7} & - & - & 69.90 & - & - & \underline{82.50} & - & - & -\\
\midrule
\multirow{4}*{\textbf{Seq2seq-based}} & BartNER \cite{b8} & 70.08 & 71.21 & 70.64 & 82.09 & 77.42 & 79.69 & 77.20 & 83.75 & 80.34 \\
~ & MAPtr \cite{b28} & 75.50 & 71.80 & 72.40 & \textbf{87.90} & 77.20 & 80.30 & - & - & -\\
~ & Debias-DDA \cite{b9} & 71.35 & 71.86 & 71.60 & 81.09 & 78.13 & 79.58 & 77.88 & 83.77 & 80.72 \\
~ & Debias-CSL \cite{b10} & 72.33 & 71.01 & 71.66 & 81.86 & 78.48 & 80.14 & 78.68 & 83.63 & 81.01 \\
\midrule
\multirow{2}*{\textbf{LLM-related}} & LLaMA2-13B \cite{b14} & - & - & - & - & - & 76.87 & - & - & 77.54 \\
~ & LLaMA-7B \cite{b15} & - & - & 47.10 & - & - & - & - & - & - \\
\midrule
\multirow{2}*{\textbf{Others}} & SegHyper \cite{b16} & 72.10 & 48.40 & 58.00 & 83.80 & 60.40 & 70.30 & 79.10 & 70.70 & 74.70 \\
~ & TBM \cite{b17} & 68.90 & 69.00 & 69.00 & 80.50 & 75.00 & 77.70 & 78.10 & 81.20 & 79.60 \\
\midrule
\multirow{3}*{\textbf{Grid-tagging}} & MAC \cite{b11} & 70.50 & 72.50 & 71.50 & 84.30 & 78.20 & 81.20 & 78.20 & \textbf{84.70} & 81.30 \\
~ & $\text{W2NER}^\dagger$ \cite{b12} & 74.52 & 70.91 & 72.67 & 85.05 & 79.92 & 82.41 & 79.04 & 84.30 & 81.58 \\
~ & $\text{TOE}^\dagger$ \cite{b13} & \textbf{76.23} & 69.60 & 72.77 & 86.54 & 78.38 & 82.26 & 79.76 & 84.07 & \underline{81.86} \\
\midrule
~ & \textbf{GapDNER (ours)}& 74.48 & \textbf{72.83} & \textbf{73.65} & 85.52 & \textbf{80.39} & \textbf{82.87} & \textbf{81.36} & 84.13 & \textbf{82.72} \\
\bottomrule
\end{tabular}
}
\end{table*}

\subsection{Main Results}
The overall performance of our proposed GapDNER on discontinuous NER is presented in Table~\ref{main-result-disc}. We could observe that GapDNER consistently outperforms all recent baselines in each dataset. Compared with previous best results, our GapDNER achieves 0.75\%, 0.37\% and 0.86\% F1 improvements in the CADEC, ShARe13 and ShARe14 datasets, respectively, leading to new state-of-the-art results for discontinuous NER. The maximum improvement achieved on the complex dataset ShARe14 demonstrates the superior adaptability of our model to various real-world scenarios. Meanwhile, GapDNER obtains the best recall in CADEC and ShARe13, and the highest precision in ShARe14. Experimental results demonstrate the outstanding capability of the proposed model in recognizing discontinuous entities.

We attribute the performance improvements to several core factors. Different from other grid tagging and span-based methods, we novelly add the learning of context gaps, which can help locate entity fragments and alleviate decoding ambiguity of discontinuous entities. We perform the linear attention computation on internal tokens of each span and capture the structural regularity within both continuous entities and entity fragments. The application of criss-cross attention models the semantic relations of overlapped spans, further enhancing span classification performance. We finally adopt the BFS algorithm to find all possible paths, which generates more valid discontinuous entities and promotes entity recall.

\subsection{Ablation Study}
We conduct necessary ablation experiments on three datasets to demonstrate the effectiveness of each component. As presented in Table~\ref{ablation-study}, both the biaffine and linear attention modules have a significant impact on the model. Removing either of them leads to an obvious decline in performance, which indicates the intra-span regularity module captures typical classification features. Meanwhile, deleting the criss-cross attention module results in a noticeable drop in F1 score, revealing the learning of inter-span relations can further enhance span classification. In addition, the application of BiLSTM also makes a slight contribution to F1 performance.

\begin{table}
\caption{\label{ablation-study}
Model ablation studies with span-level F1 scores. \textbf{Bold} and \underline{underline} indicate the largest and second largest decreases, respectively. ``w/o Biaffine'' denotes replacing Biaffine with concatenating head and tail representations.
}
\centering
\resizebox{\linewidth}{!}{ 
\begingroup
\renewcommand{\arraystretch}{0.8} 
\begin{tabular}{llll}
\toprule
Model & CADEC & ShARe13 & ShARe14 \\
\midrule
\textbf{GapDNER} & \textbf{73.65} & \textbf{82.87} & \textbf{82.72}\\
\midrule
w/o BiLSTM & 73.47 (0.18$\downarrow$) & 82.63 (0.24$\downarrow$) & 82.44 (0.28$\downarrow$) \\
w/o Biaffine & 71.85 (\underline{1.80}$\downarrow$)  & 80.92 (\textbf{1.95}$\downarrow$) & 81.30 (\underline{1.42}$\downarrow$) \\
w/o Linear Attn & 71.68 (\textbf{1.97}$\downarrow$) & 81.35 (\underline{1.52}$\downarrow$) & 81.13 (\textbf{1.59}$\downarrow$) \\
w/o Criss-cross Attn & 72.53 (1.12$\downarrow$) & 82.11 (0.76$\downarrow$) & 81.79 (0.93$\downarrow$)\\
\bottomrule
\end{tabular}
\endgroup
}
\vspace{-3.1mm}
\end{table}

\subsection{Further Analysis}
\begin{figure}[htbp]
\centerline{\includegraphics[width=\linewidth]{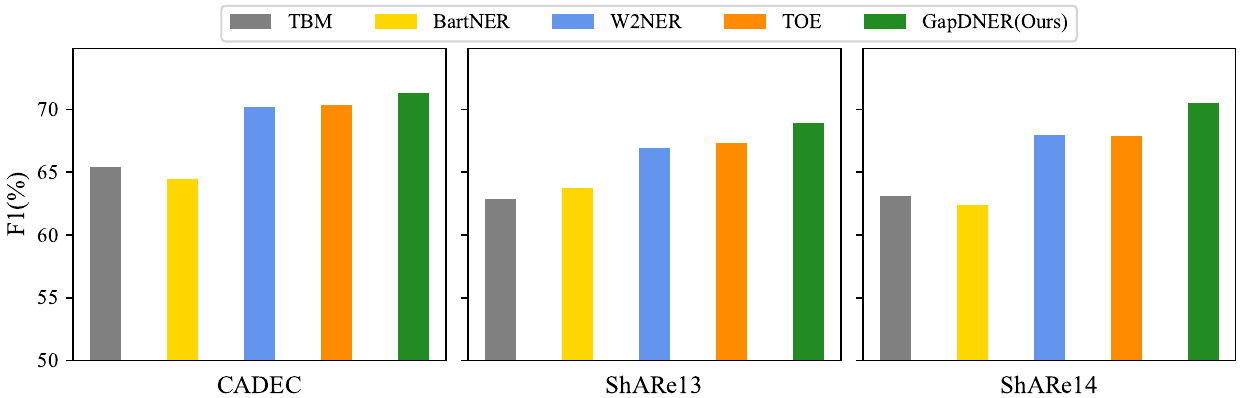}}
\caption{Results comparisons on sentences containing discontinuous entities.}
\label{fig5}
\vspace{-3mm}
\end{figure}

\textbf{Effectiveness on Discontinuous Entities} As mentioned above, discontinuous entities account for about 10\% of all entities in each data set. The evaluation results on total entities may not sufficiently show the strengths of our model in recognizing discontinuous mentions. In this case, we experiment on the subset of original test data where each sentence contains at least one discontinuous entity. From Fig.~\ref{fig5}, we find that grid-tagging models significantly outperform seq2seq-based and other methods in identifying discontinuous mentions. More importantly, our GapDNER achieves the highest F1 score compared with grid-tagging and other baselines, demonstrating its effectiveness on discontinuous entity identification.

\begin{figure}[htbp]
\centerline{\includegraphics[width=\linewidth]{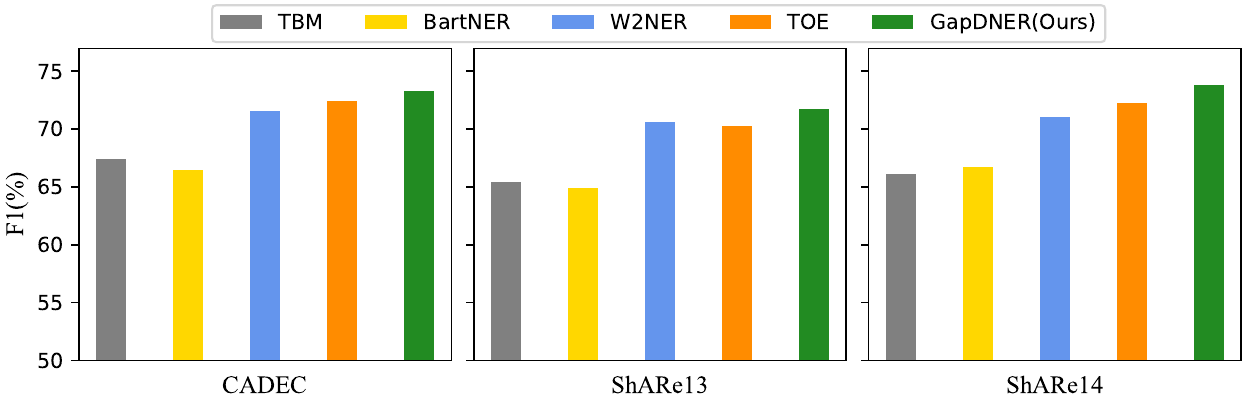}}
\caption{Results comparisons on sentences having overlapped entities.}
\label{fig6}
\end{figure}

\textbf{Performance on Overlapped Entities} We observe that overlapped structures always appear together with discontinuous entities in three biomedical data sets. To further explore the ability of our model to recognize overlapped structures, we conduct analysis experiments on the test data subset where each sentence has at least one pair of overlapped entities. As shown in Fig.~\ref{fig6}, our model still achieves the optimal results on each dataset, and shows the most significant improvement on ShARe14, demonstrating its superior ability to handle complex overlapped situations. Moreover, all types of methods perform better on overlapped NER than on discontinuous NER, indicating the former are less challenging than the latter.

\begin{figure}[htbp]
\centerline{\includegraphics[width=\linewidth]{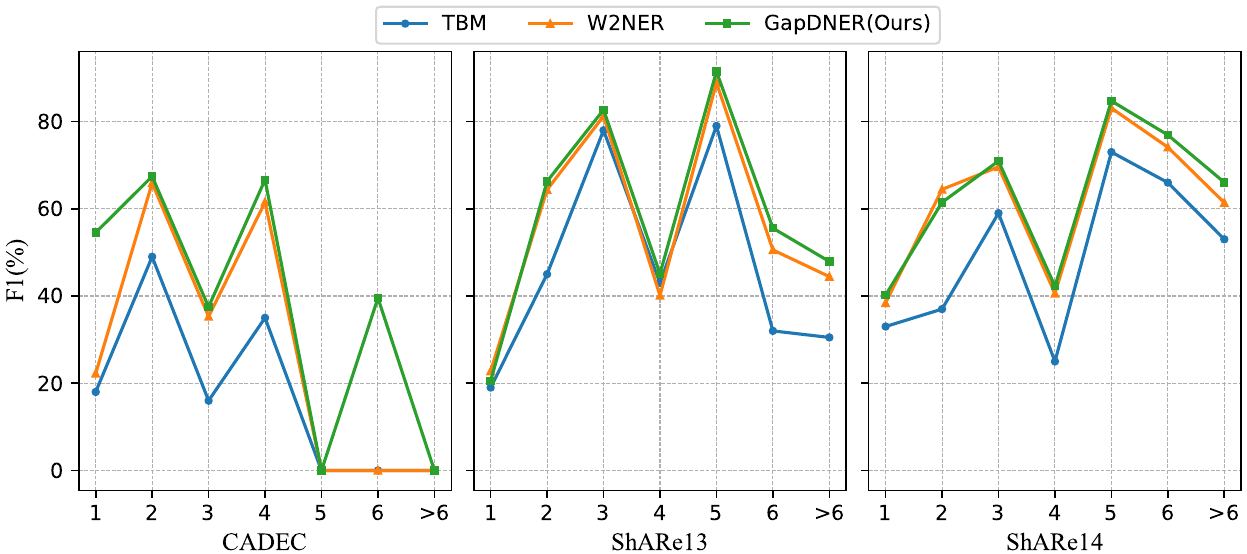}}
\caption{F1 performance of different gap lengths within discontinuous entities.}
\label{fig7}
\vspace{-3mm}
\end{figure}

\textbf{Impact of Gap Length} Longer lengths of context gaps will likely pose more challenges to discontinuous entity recognition. To verify the advantages of our gap-aware model in diverse settings, we further analyze the experimental results of test data on different gap lengths. From Fig.~\ref{fig7}, we discover that as the gap length increases, the F1 curves of three methods show similar changing trends and our model performs best in most cases, especially for long gaps. Meanwhile, each curve performs well at two smaller gap lengths (e.g., 2 and 4 for CADEC) and shows a noticeable decline in performance when the gap length is greater than 5. Interestingly, each method performs poorly at the smallest gap length 1, which may be because discontinuous entities are easily mistaken for continuous ones by the model when they have only one word to separate neighboring fragments. 

\begin{figure}[htbp]
\centerline{\includegraphics[width=\linewidth]{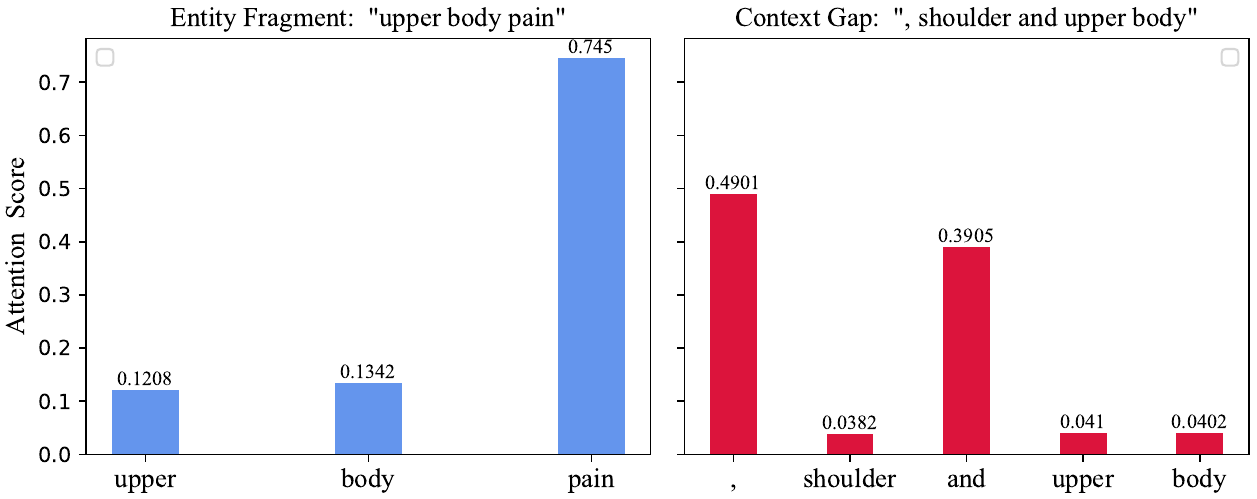}}
\caption{Linear attention distributions of the internal tokens within an entity fragment and a context gap.}
\label{fig8}
\vspace{-3mm}
\end{figure}

\begin{figure}[htbp]
\centerline{\includegraphics[width=\linewidth]{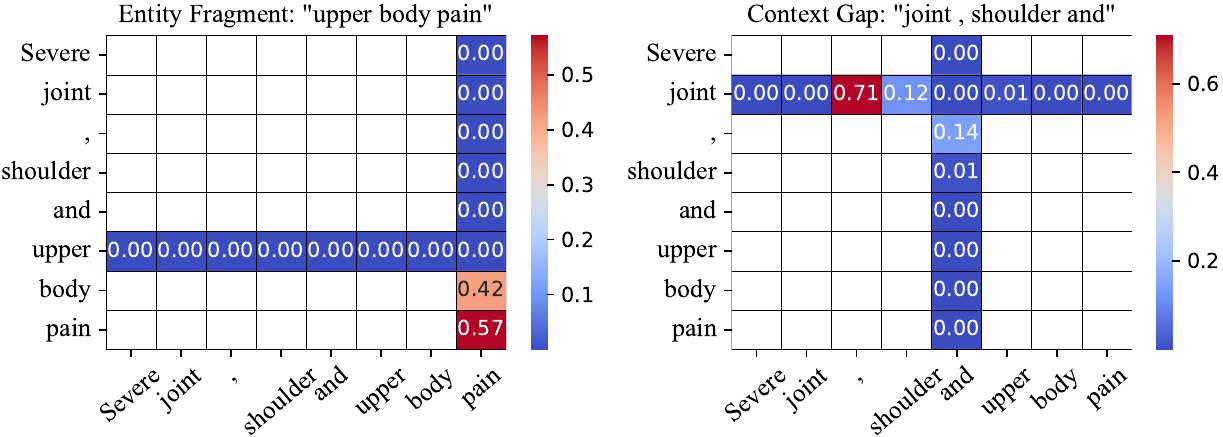}}
\caption{Criss-cross attention distributions of an entity fragment and a context gap over the token-pair feature grid, where the blank cells are not involved in the attention computation.}
\label{fig9}
\vspace{-3mm}
\end{figure}

\textbf{Case Study for Linear and Criss-cross Attention Distributions} To demonstrate the intra- and inter-span modules capture useful features for span classification, we select some cases and visualize the linear and criss-cross attention distributions in Fig.~\ref{fig8} and Fig.~\ref{fig9}, respectively. As depicted in Fig.~\ref{fig8}, we observe that the linear attention module assigns high attention scores to the core words within a type-specific span, such as ``\emph{pain}'' in the entity fragment, and the non-entity words ``\emph{,}'' and ``\emph{and}'' in the context gap. This result indicates the linear attention successfully extracts the internal composition regularity of each span. From Fig.~\ref{fig9}, we discover the fragment ``\emph{pain}'', which is in the same column as the target fragment ``\emph{upper body pain}'', is assigned the highest attention score. Meanwhile, the gap ``\emph{joint ,}'' located in the same row as the target gap ``\emph{joint , shoulder and}'', obtains the highest score in the right part. In addition, several adjacent cells of the target spans also receive notable scores. These observations confirm that the criss-cross attention module effectively models the semantic relations among overlapped spans.

\section{Conclusion}
In this paper, we propose an effective gap-aware grid tagging model for discontinuous NER, which novelly introduces context gap information to enhance entity fragment classification and resolve ambiguities in entity decoding. To accurately locate and classify spans, we develop two modules to acquire span features from intra- and inter-span aspects. The intra-span regularity extraction module comprehensively obtains span boundary interactions and internal composition regularities with the biaffine decoder and linear attention computation, respectively. The inter-span relation enhancement module adopts the criss-cross attention to model semantic relations between different spans and further promote span classifications. Experimental results on three biomedical datasets demonstrate that our model obtains new state-of-the-art performance on discontinuous NER and exhibits strong capabilities in recognizing complex entity structures.

\section*{Acknowledgment}
This work is supported by the National Key Research and Development Program of China (No.2024YFB3309702), the National Nature Science Foundation of China (No.62021002), Tsinghua BNRist and Beijing Key Laboratory of Industrial Big Data System and Application.

\vspace{12pt}

\end{document}